\pdfoutput=1

\documentclass[11pt]{article}

\usepackage[final]{emnlp2022}

\usepackage{times}
\usepackage{latexsym}
\usepackage{tabularx}
\usepackage[T1]{fontenc}

\usepackage[utf8]{inputenc}

\usepackage{microtype}

\usepackage{kotex}
\usepackage{booktabs}
\usepackage{adjustbox}
\usepackage{graphicx}
\usepackage{amssymb}
\usepackage{amsmath}

\usepackage{multirow}

\usepackage{comment}
\usepackage{subcaption}
\usepackage{makecell}
\usepackage{xcolor}

\title{Evaluating the Knowledge Dependency of Questions}
\author{
\quad Hyeongdon Moon\thanks{\quad Equal Contribution.}~~\thanks{\quad Corresponding authors. Work was done while all corresponding authors were at Riiid AI Research. Jamin Shin is currently at NAVER AI Lab.}~~$^{1}$
    \quad Yoonseok Yang\footnotemark[1]~~\footnotemark[2]~~$^{2}$
\quad \textbf{Jamin Shin}\footnotemark[2] \\
\quad \textbf{Hangyeol Yu}$^{1}$
\quad \textbf{Seunghyun Lee}$^{1}$ 
\quad \textbf{Myeongho Jeong}$^{1}$ \\
\quad \textbf{Juneyoung Park}$^{1}$
\quad \textbf{Minsam Kim}\footnotemark[2]~~$^{1}$
\quad \textbf{Seungtaek Choi}\footnotemark[2]~~$^{1}$\\
$^1$Riiid AI Research\\
$^2$UC Berkeley \\
\texttt{hyeongdon.mun@riiid.co},
\texttt{yoonseok@berkeley.edu},\\
\texttt{jayshin.nlp@gmail.com}, 
\texttt{\{minsam.kim, seungtaek.choi\}@riiid.co}\\
}

\begin{document}
\maketitle


\newcommand{\todoc}[2]{{\textcolor{#1}{\textbf{#2}}}}
\newcommand{\todoblue}[1]{\todoc{blue}{\textbf{#1}}}
\newcommand{\todored}[1]{\todoc{red}{#1}}
\newcommand{\hist}[1]{\todored{\textbf{HIST:} #1}}
\newcommand{\jay}[1]{\todoblue{\textbf{JAY:} #1}}
\newcommand{\doni}[1]{\todoc{green}{\textbf{DONI:} #1}}
\newcommand{\yoon}[1]{\todo[inline,color=blue!20!white]{\textbf{YOON:} #1}}

\newcommand{\ours}{$\mathbf{KDA}$}
\newcommand{\oursstar}{$\mathbf{KDA_*}$}
\newcommand{\ourssoft}{$\mathbf{KDA_{cont}}$}
\newcommand{\oursdisc}{$\mathbf{KDA_{disc}}$}
\newcommand{\ourshuman}{$\mathbf{KDA_{human}}$}

\newcommand{\ourssmall}{$\mathbf{KDA_{small}}$}
\newcommand{\ourslarge}{$\mathbf{KDA_{large}}$}

\begin{abstract}
The automatic generation of Multiple Choice Questions (MCQ) has the potential to reduce the time educators spend on student assessment significantly.
However, existing evaluation metrics for MCQ generation, such as BLEU, ROUGE, and METEOR, focus on the n-gram based similarity of the generated MCQ to the gold sample in the dataset and disregard their educational value.
They fail to evaluate the MCQ's ability to assess the student's knowledge of the corresponding target fact.
To tackle this issue, we propose a novel automatic evaluation metric, coined \textbf{Knowledge Dependent Answerability (\ours)}, which measures the MCQ's answerability given knowledge of the target fact.
Specifically, we first show how to measure \ours~based on student responses from a human survey.
Then, we propose two automatic evaluation metrics, \oursdisc~and \ourssoft, that approximate \ours~by leveraging pre-trained language models to imitate students' problem-solving behavior.
Through our human studies, we show that \oursdisc~and \ourssoft~have
strong correlations with both (1) \ours~and (2) usability in an actual classroom setting, labeled by experts. 
Furthermore, when combined with n-gram based similarity metrics, \oursdisc~and \ourssoft~are shown to have a strong predictive power for various expert-labeled MCQ quality measures.
\footnote{\quad Our code is released here: \url{https://github.com/riiid/question-score}}
\end{abstract}
\section{Introduction}

\begin{figure}[t!]
\centering
    \begin{subfigure}[b]{0.97\columnwidth}
        \centering
        \includegraphics[clip,width=\columnwidth]{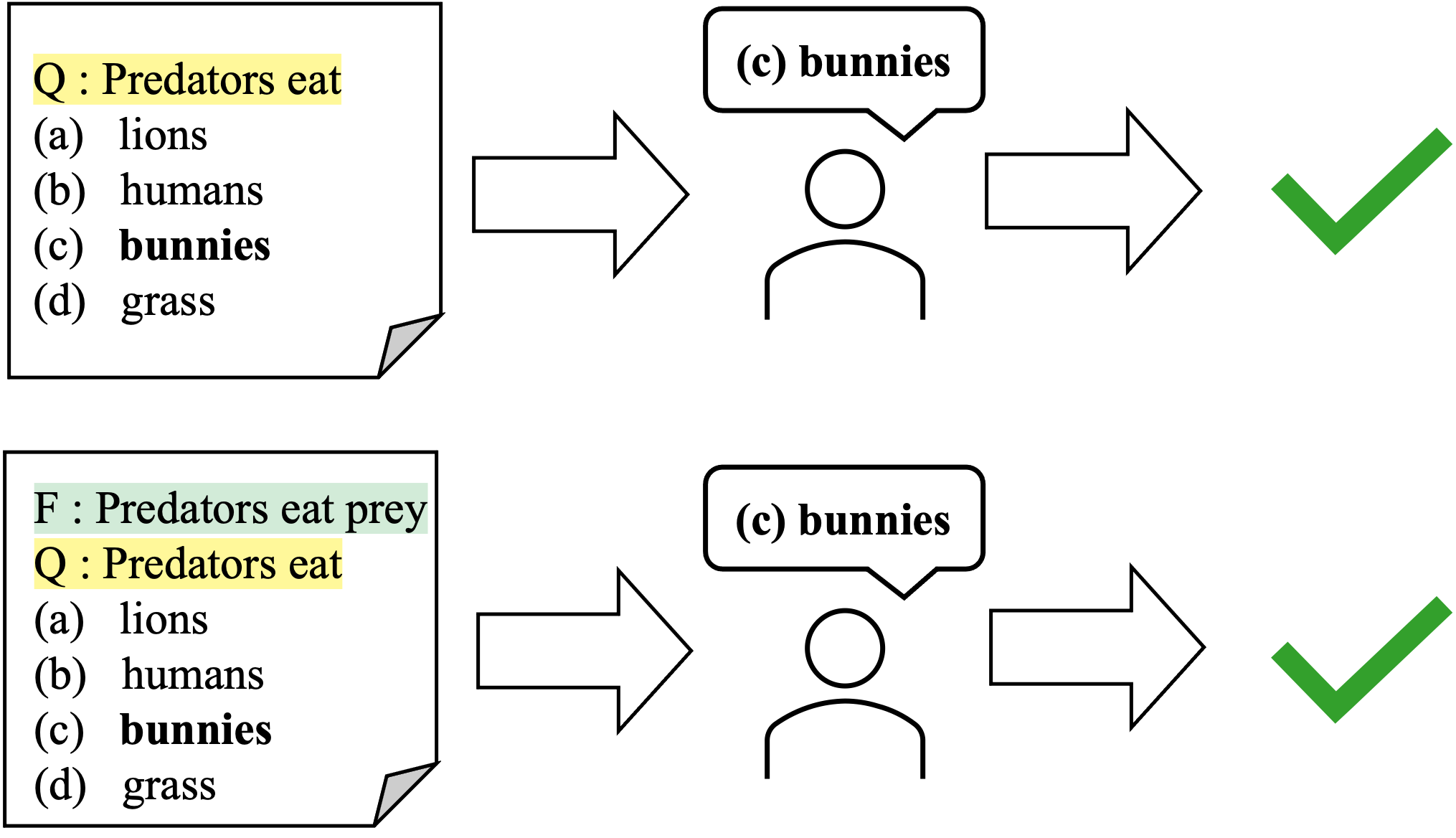}
        \caption{Bad question (too easy)}
        \label{fig:}
    \end{subfigure}
    
    \begin{subfigure}[b]{\columnwidth}
        \centering
        \includegraphics[clip,width=\columnwidth]{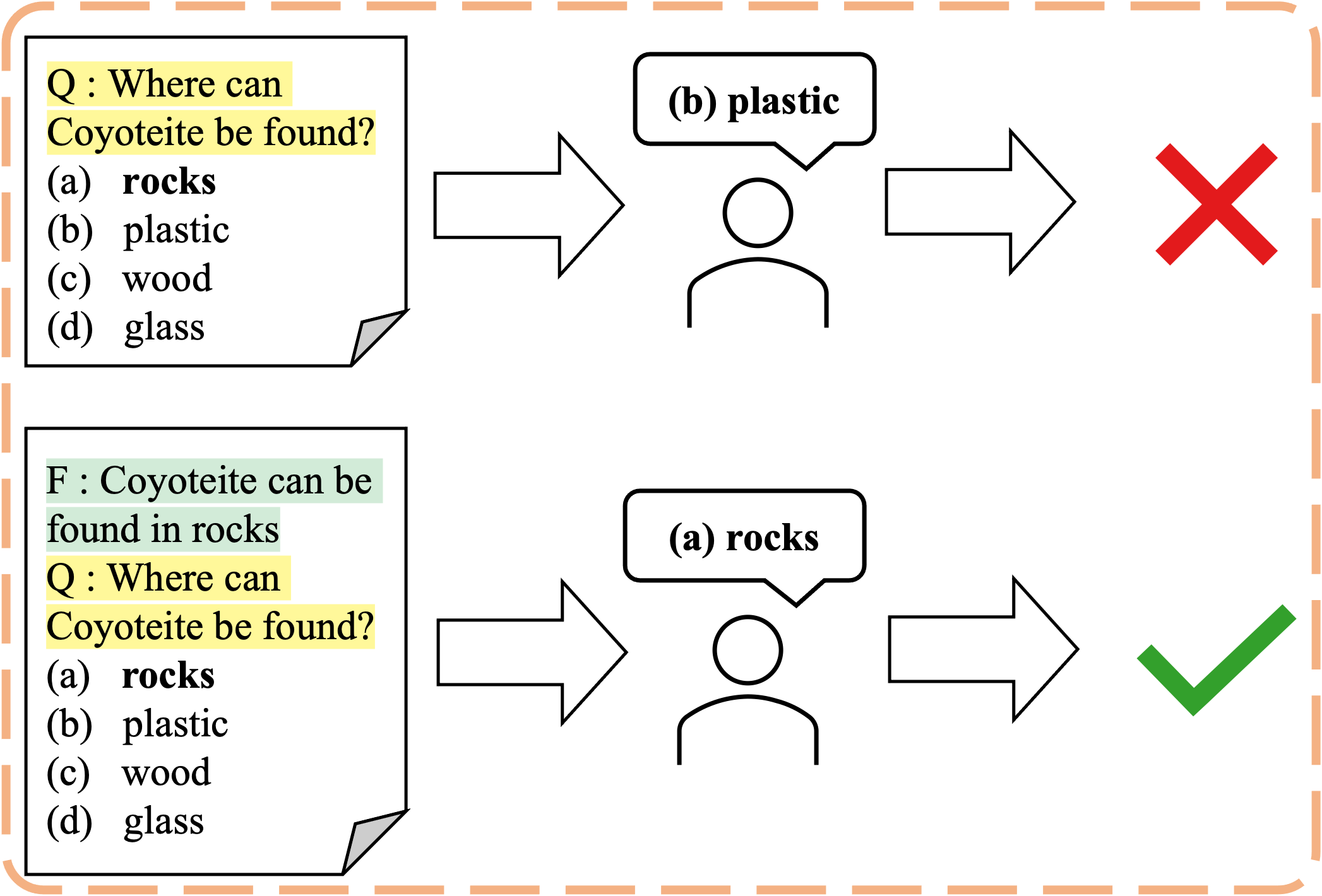}
        \caption{Good question}
        \label{fig:good-question}
    \end{subfigure}
    
    \caption{Problem formulation. 
    A multiple-choice question designed to test the knowledge of a target fact must satisfy the following criterion: a question has to be answered by only a student who knows the fact, not a student who doesn't.
    }
  \label{fig:intro}
\end{figure}

Multiple-Choice Question (MCQ), comprised of a question stem, an answer, and a set of distractors, is one of the most widely used student assessment tools. Since manually generating exam-style questions is a complex, labor-intensive process that requires training, experience, and resources, automatic question generation (AQG) techniques were introduced \cite{kurdi2020systematic}. If automatic MCQ generation methods develop to a level that requires only minor adjustments by educators, it can meet the real-world demand for creating a massive amount of MCQ sets in seconds. 

Despite the importance of AQG in educational purposes, the task did not receive much attention and was not actively introduced in the education field due to the limitation of evaluation methods. 
Previous AQG works mostly evaluate their methods based on how similar the generated questions/distractors are to the gold questions/distractors, using n-gram based similarity metrics such as BLEU \citep{bleu}, ROUGE \citep{rouge}, METEOR \citep{meteor}, or BERTScore \citep{bertscore}. 
However, all metrics mentioned above share the following limitations: 
(1) reliability of the evaluation depends on the quality of the reference dataset against which similarity is measured, and  
(2) n-gram-based similarity of a question does not directly measure the usability of the question as an assessment tool.
Though a prior work \citep{nema2018towards} proposed an alternative metric of answerability, the suggested metric does not account for whether the question's answerability crucially depends on knowledge of the target fact and also requires reference made by humans. AQG will be applied in the real-world much more easily if we address these limitations.

To this end, this paper suggests a new evaluation criterion for MCQ generation, coined \textbf{K}nowledge \textbf{D}ependent \textbf{A}nswerability (\ours).
\ours~measures whether a real student who got the problem wrong can choose the correct answer after gaining the knowledge the problem wants to test. 
\ours~is based on the rationale that a well-designed MCQ should be answerable if the student knows the target fact being tested, which is briefly described in Figure \ref{fig:intro}. 

However, as such human-dependent measurement is hard to be applied in real world scenario, we propose two automatic variants, \ourssoft~and \oursdisc. 
To automate the measurement of \ours, we regard the Pre-trained Language Models (PLMs) as question solvers that imitate students, based on the fact that large PLM's have strong question-answering abilities \citep{qasurvey, roberts2020much}.
To show the validity of the metrics, we ask two research questions: RQ1) When replacing students with PLMs, do the automated metrics (\ourssoft~and \oursdisc) behave similarly to \ours~? and RQ2) Can we use MCQs with high \ourssoft~and \oursdisc~ in education field meaningfully?

To answer these research questions, we asked 116 students to solve 480 MCQs, which include both automatically generated questions by various methodologies and gold questions from three MCQ datasets. 
Among them, 96 questions were randomly selected, and a qualitative survey was conducted with experts to see if they were suitable for educational use. 
Through our experiments, we demonstrate that \oursdisc~and \ourssoft~both have a strong correlation not only with \ours~as measured based on student responses but also with expert Likert score that measures the MCQ's overall usability in the classroom setting. 

Our main contributions are as follows:
\begin{itemize}
    \item We propose a novel, reference-free metric Knowledge-Dependent Answerability (\ours) that evaluates given MCQ's value as an assessment tool, and two automatic alternatives (\oursdisc, \ourssoft) of approximating \ours with PLMs. 
    \item We validate the usability of \oursdisc~and \ourssoft~through extensive human studies.
    \item We release our code for \ours, in order to facilitate usage in public domain.
\end{itemize}

\section{Related Work}
\subsection{Multiple-Choice Question Generation}

Multiple choice question (MCQ) is comprised of a question, an answer, and a set of distractors. MCQ makes student assessment feasible, as it disambiguates a question by providing options to choose from  \citep{rachmat2019use}.
Especially, there have been multiple findings 
that active learning through question answering helps increasing learning gain of students \citep{crouch2001peer, koedinger1997intelligent, wang2022towards}.
However, instructors suffer from generating high quality MCQs due to their limited resources.

Accordingly, there have been various attempts at automating MCQ generation. 
Here, we briefly review the automatic MCQ generation with regards to two important components: (1) question generation and (2) distractor generation. 

\paragraph{Question Generation} 
Question Generation (QG) is usually formulated as a task where an appropriate question must be generated given a reference document \citep{qa-qg}, where generated question can be answered from the reference document. 
For MCQ generation, target answer is usually given with the reference document, which serves as the answer for the generated question \citep{leaf}.
With the advent of deep learning, various neural networks have been used for QG: LSTM-based \citep{lstm-qg} and transformer-based \cite{quizgen, qg-rewards}.

\paragraph{Distractor Generation} 
The goal of Distractor Generation (DG) is to receive a reference document, a question, and a target answer as inputs, and then output a set of distractors. 
There are mainly two classes of DG methods, namely, knowledge-based DG and language model-based DG. 

Here, we regard knowledge-based DG as the superset of ontology-based DG and Knowledge-Driven DG. Historically, ontology-based DG was first proposed, which retrieves distractors that are similar to the answer according to the given domain-specific ontology \cite{ontologyMCQ}. 
More recently, Knowledge-Driven DG has been proposed (\citet{kddg}; KDDG), which uses a general-purpose knowledge base to select a pool of distractors, then rank the distractors using a feature-based model. In general, while knowledge-based DG provides plausible distractors semantically similar to the answer, they heavily rely on the quality of the knowledge base, and offer limited scalability.

To overcome the limitations of knowledge-based DG, recent works leveraged PLMs for DG \citep{bdg, leaf}.
Specifically, by fine-tuning a PLM to replicate gold distractors given the corresponding question and answer, the PLM learns how to make sensible distractors without relying on a knowledge base.

In this work, we employ T5~\cite{leaf} for both question generation and distractor generation due to its superior performance. 
Further, we implement KDDG and evaluate its performance using our proposed evaluation metric to represent knowledge-based DG.

\subsection{Automatic Evaluation of MCQs}
Following most works in language generation, prior works in QG and DG rely on n-gram similarity-based metrics such as BLEU \citep{bleu}, ROUGE \citep{rouge}, and METEOR \citep{meteor} for automatic evaluation.  
BERTscore \citep{bertscore} improves upon these metrics by performing n-gram matching with BERT-based contextualized embeddings.  
However, all of the aforementioned metrics only consider how similar the generated output is to the gold samples \citep{nema2018towards}, and fail to consider the value of generated questions as a student assessment tool.


To address this issue, \cite{nema2018towards} considers Answerability of the question. 
In particular, they first collect human scores on whether an MCQ is answerable, then suggests four features for predicting the answerability score: relevant words, named entities, question words, and function words.
Answerability metric is then defined as a learned weighted sum of the above features extracted from human, showing improved correlation with human judgement upon BLEU. 
Q-BLEU, which combines Answerability and BLEU, is shown to improve upon BLEU in terms of correlation with human judgment. 
However, this approach requires human annotations, and it is unlikely that results in one domain will extend to another. Also, the suggested answerability does not account for the source text given with the MCQ. \cite{wang2022towards} points out the low adoption of QG Systems in classrooms, requesting the QG system researchers to focus on educational needs.

\begin{figure*}[ht!]
\centering
  \includegraphics[width=1.0\textwidth]{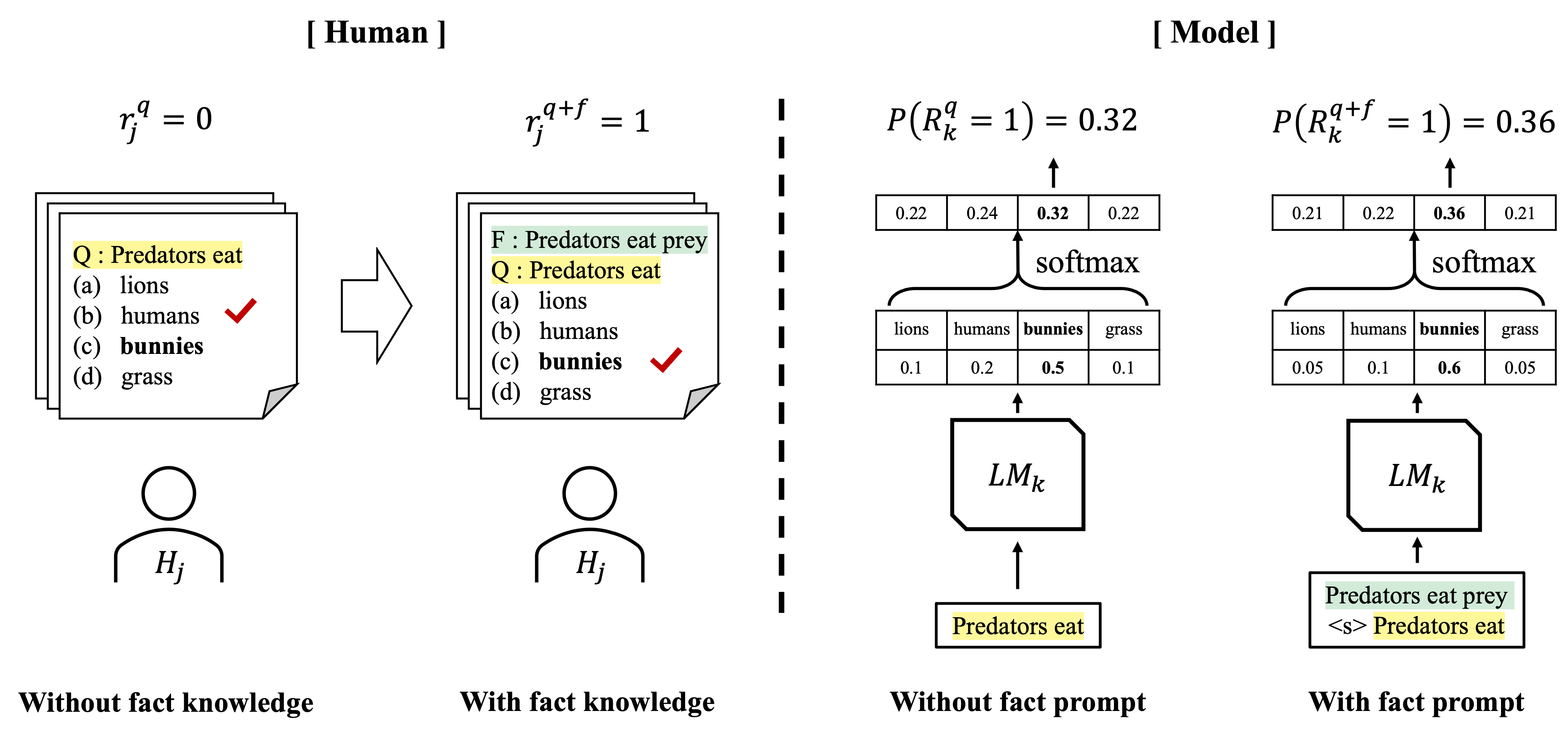}
    \caption{ Flow diagram of human experiment and model inference. Left: human response phase and example. Each phase of human experiment tests 40 facts. After solving all questions without fact, each student solves all question again with corresponding fact. Right: inference of the language model. For several types of MCQ solver, we inference twice — with, and without the target fact.}
  \label{fig:main_figure}
\end{figure*}

\section{Knowledge Dependent Answerability}
\label{sec:approach}

To properly evaluate the question's answerability, we propose to measure \textbf{Knowledge Dependent Answerability (\ours)}. 
In particular, for each MCQ, we measure the proportion of students who answer the question correctly when they know the target fact being tested. 
Our rationale is that a good MCQ's answerability must crucially depend on the knowledge of the target fact. 
Then, in order to automatically measure KDA without human trials, we replace the students with Pre-trained Language Models based on their question-answering abilities \citep{qasurvey,roberts2020much}.

\subsection{Measuring \ours~with student responses}
Our goal is to measure the probability that a student will answer the MCQ correctly given that they know the target fact being tested. 
For this, we consider the participants who initially answer the target fact's pair question incorrectly as the ones who do not know the target fact. Then, we count the subset of those participants who answer the question correctly after the fact is shown. 
In other words, we measure the following:
\newcommand\given[1][]{\:#1\vert\:}
\begin{align} 
\label{eq:KDA}
    KDA(q) &= P(R^{q_+f} = 1 \given R^{q} = 0) \\
    &\approx \frac{\sum_{j}{(1-r^q_{j})r^{q+f}_{j}}}{\sum_{j}{(1-r^q_{j})}}, \nonumber
\end{align}
where $R^{q}$ and $R^{q+f}$ represent binary random variables for answer correctness before and after showing the fact, respectively.  
Here, lower-case variables represent sampled values, and $j$ runs over all samples collected from human participants. 





\subsection{Measuring \ours~with Language Models}
In order to approximate \ours~(Equation~\ref{eq:KDA}), we use PLMs to emulate problem-solving students.
In particular, in order to obtain $r^{q}$, we let the $j$-th language model to predict the answer given question $q$, as shown in Figure~\ref{fig:main_figure}. 
We obtain $r^{q+f}_{j}$ similarly, except we prompt the target fact in front of the question, as shown in Figure~\ref{fig:main_figure}.  
We propose two versions of PLM-based \ours, namely, \oursdisc~and \ourssoft~(disc for \textit{discrete}, and cont for \textit{continuous}). 

\begin{equation} 
\begin{split} \label{eq:recall_2}
    \mathbf{KDA_{disc}}(q) = \frac{\sum_{j}{(1-r^q_{j})r^{q+f}_{j}}}{\sum_{j}{(1-r^q_{j})}}.
\end{split}
\end{equation}

\oursdisc~exactly replicates Equation~\ref{eq:KDA}, but the binary values $r_{j}^{q}$ and $r_{j}^{q+f}$ are obtained from various PLM outputs instead of real students. Specifically, when the logit for the correct answer is the largest among the options, $r_{j}^{q}$ (or $r_{j}^{q+f}$) equals $1$, and otherwise $0$.


\begin{equation} 
\begin{split} \label{eq:recall_3}
    \mathbf{KDA_{cont}}(q) = \frac{\sum_{j}{P(R_{j}^q = 0)P(R_{j}^{q+f} = 1)}}{\sum_{j}{P(R_{j}^q = 0)}},
\end{split}
\end{equation}

With language models, we can utilize probability outputs which may contain richer information compared to discretized values. Thus, we further propose the continuous version by replacing the binary values of Equation~\ref{eq:KDA} with probability outputs from the language models. \ourssoft~is interpreted as a weighted average of correctness probability $P(R_{j}^{q+f} = 1)$ across models, weighted by each model's probability of incorrect response without being shown the target fact. \ourssoft~ is also more stable than \oursdisc, as the denominator of the Equation \ref{eq:recall_2} can be zero if all model answer correctly to the question without fact.

\section{Experiments}
In this section, we first discuss the settings of our experiments to verify the efficacy of \ours, and then describe student and expert studies we conducted in detail. Lastly, we show our results for respective experiments. 
\label{sec: experiment}
\subsection{Preliminaries}
We designed our experiments to answer the following research questions:

\begin{itemize}
  \setlength\itemsep{-4.0pt}
  \setlength\topsep{-4.0pt}
   \item \textbf{RQ1}: Whether we can automatically measure \ours~ with \oursdisc~ and \ourssoft~ (Section~\ref{subsection:corr})
   \item \textbf{RQ2}: Whether \ours, \oursdisc, and \ourssoft~ correlate well with the judgments of real-world educators (Section~\ref{subsection:teacher})
\end{itemize}
\begin{table*}[]
\begin{tabular}{@{}llll@{}}
\toprule
            & OBQA                                                           & TabMCQ                                                                                          & SciQ                                                                                                                      \\ \midrule
Fact        & predators eat prey                                             & \begin{tabular}[c]{@{}l@{}}Urban sprawl creates\\ thermal pollution\end{tabular}                & \begin{tabular}[c]{@{}l@{}}Plant hormones are chemical signals\\ that control different processes in plants.\end{tabular} \\ \midrule
Question    & Predators eat                                                  & \begin{tabular}[c]{@{}l@{}}What type of pollution\\ does Urban sprawl create?\end{tabular}      & \begin{tabular}[c]{@{}l@{}}What chemical signals in plants\\ control different processes?\end{tabular}                    \\ \midrule
Answer      & bunnies                                                        & thermal pollution                                                                               & plant hormones                                                                                                            \\ \midrule
Distractors & \begin{tabular}[c]{@{}l@{}}lions\\ humans\\ grass\end{tabular} & \begin{tabular}[c]{@{}l@{}}air pollution\\ radioactive pollution\\ noise pollution\end{tabular} & \begin{tabular}[c]{@{}l@{}}produce hormones\\ nitrogen hormones\\ Human Hormones\end{tabular}   \\ \bottomrule                         
\end{tabular}
\caption{Example questions from each dataset}
\label{tab: example-q}
\end{table*}
\noindent
To answer these questions, we need MCQ Generator Models to create questions on several datasets and MCQ Solver Language Models to measure \oursdisc~and \ourssoft.

\paragraph{MCQ Generator Models}
We divide MCQ Generation as only generating the distractors, and generating both distrators and question stem. 3 types of MCQ Generator models were used: two of them are of Distractor-only generation models and one generates both the question stem and distractors.

We first discuss the Distractor-only Generation models, which use fixed question stems provided from the datasets.
For knowledge-based DG, we implemented KDDG \citep{kddg}, which picks distractor candidates from the knowledge-base according to the topic distribution modeled by LDA \citep{blei2003latent} and then ranks them using learned features such as similarity to the answer. 
Following the authors' implementation, we selected Probase~\citep{wu2012probase} as the knowledge base, and pair-wise LambdaMART ranker~\citep{burges2011learning} as the ranker.
As for PLM-based DG, we fine-tuned the \texttt{T5-Large model}~\citep{raffel2020exploring} on the RACE dataset \citep{race}, a large-scale reading comprehension MCQ dataset. 
To be specific, following~\citep{leaf}, we provided the question, answer, and the reference document as an input, then obtained three distractors.
As shown in Table ~\ref{tab: dg-bleu}, the first distractors scored 46.59 BLEU1 on test data. 
We refer to this model as \textbf{T5DG}.

For \textit{Question Generation (QG)}, we fine-tuned \texttt{T5-Large}~\cite{raffel2020exploring} models on each datasets to generate the question stems. The distractors are then generated with the above \textbf{T5DG}. This model, which we refer to as \textbf{QDG}, scored BLEU1 of 19.4, 53.3, 65.9 in OBQA, TabMCQ, and SciQ, respectively.

We prepared and evaluated MCQs generated by four different MCQ generation pipelines, as shown in Table~\ref{tab: baseline-models}. 
First, we evaluated MCQs generated by KDDG and T5DG based on human-created questions. 
Also, we evaluated MCQs whose question stem and distractors are both model-generated. 
Finally, we evaluated MCQs fully generated by humans, i.e., both questions and distractors are human-generated.  
More training details and results are included in the Appendix.

\begin{table}[t!]
\centering
\begin{tabular}{@{}lcc@{}}
\toprule
Generator    & Distractors & Question Stem \\ \midrule
Human   & Human      & Human    \\
KDDG   & KDDG      & Human    \\
T5DG   & T5-DG      & Human    \\
QDG & T5-DG     & T5-QG     \\ \bottomrule
\end{tabular}
\caption{Baseline Models for QG/DG}
\label{tab: baseline-models}
\end{table}

\begin{table}[t]
\begin{tabular}{l|llll}
\toprule
   & BLEU1 & BLEU2 & BLEU3 & BLEU4 \\ \midrule
D1 & 46.59 & 38.33 & 34.31 & 32.02 \\
D2 & 25.8  & 20.08 & 17.58 & 16.15 \\
D3 & 28.33 & 23.07 & 20.73 & 19.46 \\ \bottomrule
\end{tabular}
\caption{BLEU scores for T5-DG in RACE}
\label{tab: dg-bleu}
\end{table}

\paragraph{MCQ Solver Language Models}
To calculate \oursdisc~and \ourssoft, we prepare pre-trained language models as general-purpose MCQ solvers. 
In particular, in order to emulate students with various knowledge states and reasoning capabilities, we prepared 18 different PLM's of various types and sizes. 
For T5 models, we used the ones fine-tuned for Closed Book QA~\cite{roberts2020much}.
For other models, we fine-tuned the models with RACE dataset. 
A complete list of solver models can be found in Appendix.

\paragraph{Datasets}
We used three real-world MCQ datasets for our experiments. 
Every MCQ contained in each dataset has three distractors and one answer. 

\begin{itemize}
    \setlength\itemsep{-4.0pt}
    \setlength\topsep{-4.0pt}
    \item OpenBookQA (OBQA) \citep{obqa} is comprised of 5,957 elementary-level science questions. Answering the questions requires multi-step reasoning and commonsense knowledge.
    \item TabMCQ \cite{tabMCQ} contains 9,091 crowd-sourced MCQ science questions based on fact-based relation tables.
    \item SciQ \cite{sciQ} contains 13,679 science exam questions about physics, chemistry, and biology. 
\end{itemize}

\subsection{RQ1: Correlation of \ours~and $\mathbf{KDA_*}$}
\label{subsection:corr}

\begin{table}[t!]
\begin{adjustbox}{width=1\columnwidth}
\label{tab:h5correlation}
\begin{tabular}{@{}lllll@{}}
\toprule
         & OBQA            & TabMCQ         & SciQ           & All            \\ \midrule
\ourssoft~& \textbf{0.73**} & 0.16           & 0.17           & 0.74**         \\
\oursdisc~& 0.71**          & \textbf{0.3**} & 0.05           & \textbf{0.8**} \\ \midrule
BLEU     & 0.29**          & 0.14           & 0.16           & 0.26**         \\
ROUGE-L  & 0.29**          & 0.14           & \textbf{0.18*} & 0.27**         \\
METEOR   & 0.28**          & 0.12           & 0.14           & 0.21**         \\ \bottomrule
\end{tabular}
\end{adjustbox}
\caption{Pearson Correlation between \ours~and other automatic evaluation metrics. Single (double) asterisk denotes p-value under 0.05 (0.01). Best correlation results are marked bold per dataset.}
\label{tab: corr-human}
\end{table}

\begin{figure}[]
  \includegraphics[width=0.45\textwidth]{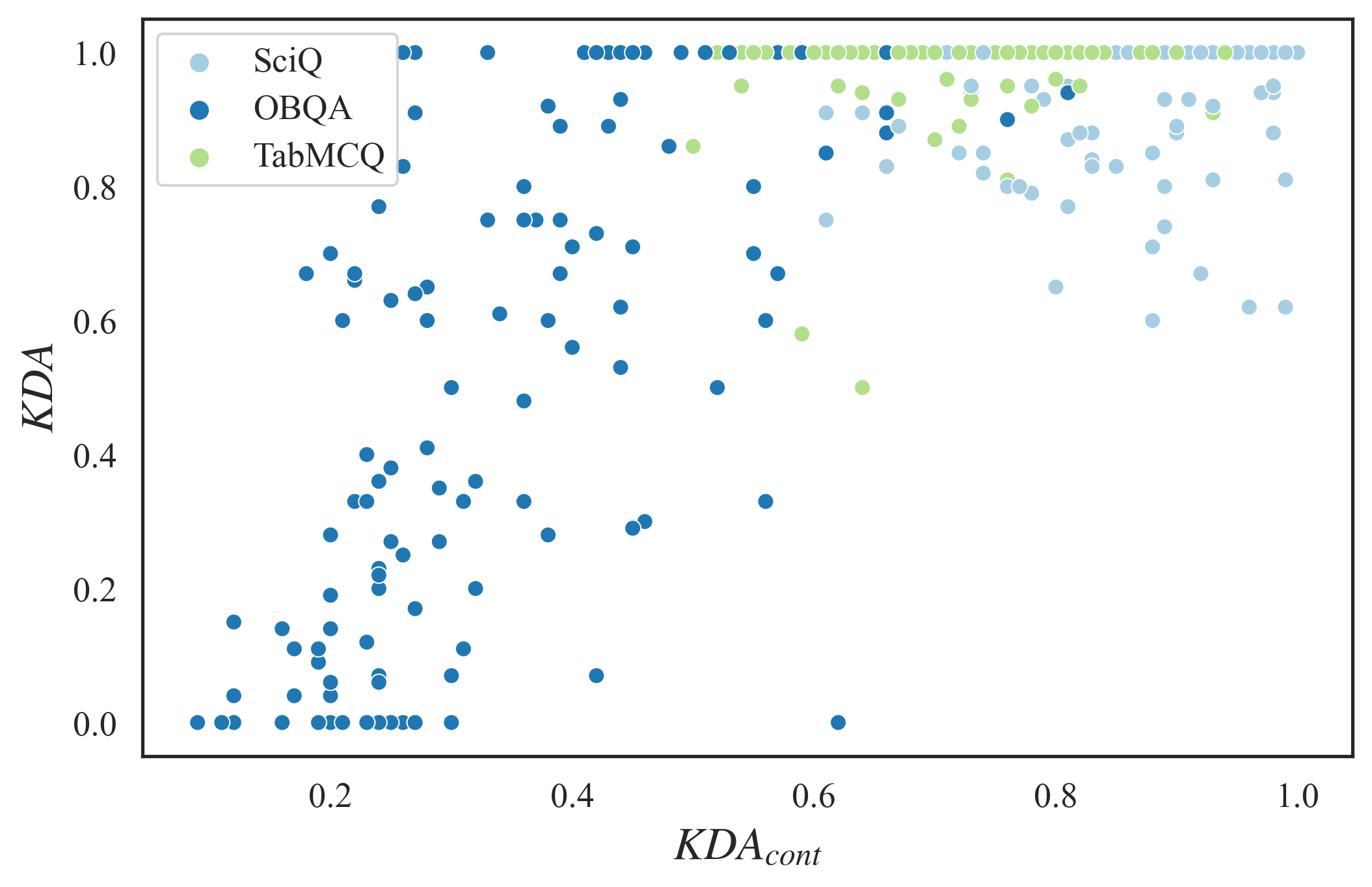}
    \caption{Scatter plot of the result. x-axis : \ourssoft, soften \ours~measured by language models. y-axis: \ours, gold knowledge dependency measured by human solvers. The overall correlation between two metrics is 0.74 }
  \label{fig:corr-human}
\end{figure}

\paragraph{Setup}
To show that automatic evaluation metrics \oursdisc~and \ourssoft~indeed have strong correlations with \ours, we conducted a large-scale human study with 116 participants. 
We randomly sampled 40 facts from each dataset (120 in total). 
Since we have four different MCQ generation pipelines for each fact, there were 4 different questions asking for the same fact. Thus, we randomly mixed the questions from different models and splitted participants into four groups. More details can be found in the Appendix.  
The survey proceeded in the following stages, where participants had to answer 120 questions for each stage:

\noindent
\textbf{1) Question solving without fact: } We first asked participants to solve the questions without showing them the relevant target facts. 

\noindent
\textbf{2) Question solving with fact: } Then, we asked participants to solve the questions with the relevant target facts given. 
\begin{table}[t!]
\centering
\begin{tabular}{@{}lll@{}}
\toprule
Size and number of LMs       & \ours & Likert \\ \midrule
LMs \textless{} 1GB (4 LMs)    & 0.65          & 0.36             \\
LMs \textless{} 1.5GB (4 LMs)  & 0.71          & 0.38             \\
LMs \textless{} 1.5GB (11 LMs) & 0.73          & 0.39             \\
All LMs (18 LMs)             & 0.74          & 0.43             \\ \bottomrule
\end{tabular}
\caption{ Pearson Correlation of \ourssoft~with \ours~and expert Likert score by the model size. We measured \ourssoft~by using different models as solvers to show performance change along the number of language models used to calculate \ourssoft. Each row of the table is obtained by using four small-size models only, the same numbers of larger-size models, 11 models under larger sizes, and the entire models. }
\label{tab: N1}
\end{table}

\begin{table*}[t!]
\begin{adjustbox}{width=1\textwidth}
\begin{tabular}{@{}cl@{}}
\toprule
\multicolumn{1}{l}{Score} & Descriptions                                                                                                   \\ \midrule
4                                & \textbf{[Strongly Agree]} This question can be readily used in the classroom.                                       \\
3                                & \textbf{[Agree]} Despite some minor flaws, I’m willing to use this question to test the fact in a classroom         \\
2                                & \textbf{[Disagree]} This question has some major flaws that needs to be revised for educational use in a classroom. \\
1                                & \textbf{[Strongly Disagree]} This question should be changed completely to be used in a classroom.                  \\ \bottomrule
\end{tabular}
\end{adjustbox}
\caption{Response types for the following question: On a scale of 1-4, how would you evaluate this question to be used in a classroom to test the fact below?}
\label{tab:response} 
\end{table*}

\paragraph{Results}
As seen in Table \ref{tab: corr-human}, both \oursdisc~and \ourssoft~show significantly higher correlation with \ours~compared to n-gram based similarity metrics such as BLEU, ROUGE, and METEOR (0.74 and 0.80 respectively, with $p < 0.01$). 
This shows that ngram-based similarity metrics do not faithfully represent the MCQ's assessment value, while \oursdisc~and \ourssoft~can provide insight into the MCQ's assessment capabilities by considering language models' problem solving behavior.

Note that correlations may vary significantly across datasets due to different dataset characteristics.
For example, although \oursdisc~and \ourssoft~correlate strongly with \ours~for OBQA, the correlation is not as strong for TabMCQ and SciQ. 
This is because questions in OBQA require multi-step reasoning and commonsense knowledge (average correctness of 0.71 after human participants are shown the facts), while TabMCQ and SciQ mostly contain easy questions that are simple paraphrases of the relevant facts (corresponding human average correctness of 0.99 and 0.96, respectively).
Therefore, while language models may find MCQs from TabMCQ and SciQ relatively easy, human participants found the MCQs too easy, leading to a relatively weak correlation. 
This can be seen in Figure \ref{fig:corr-human}, where a lot of data points for OBQA and SciQ can be found around $y=1.0$. 

In addition, Table \ref{tab: N1} shows the performance enhancement as models' size and the number of models increase. We believe a better correlation can be obtained when we use larger language models with improved reasoning capabilities to better imitate human problem solvers.

 

\begin{figure}[t]
  \includegraphics[width=0.45\textwidth]{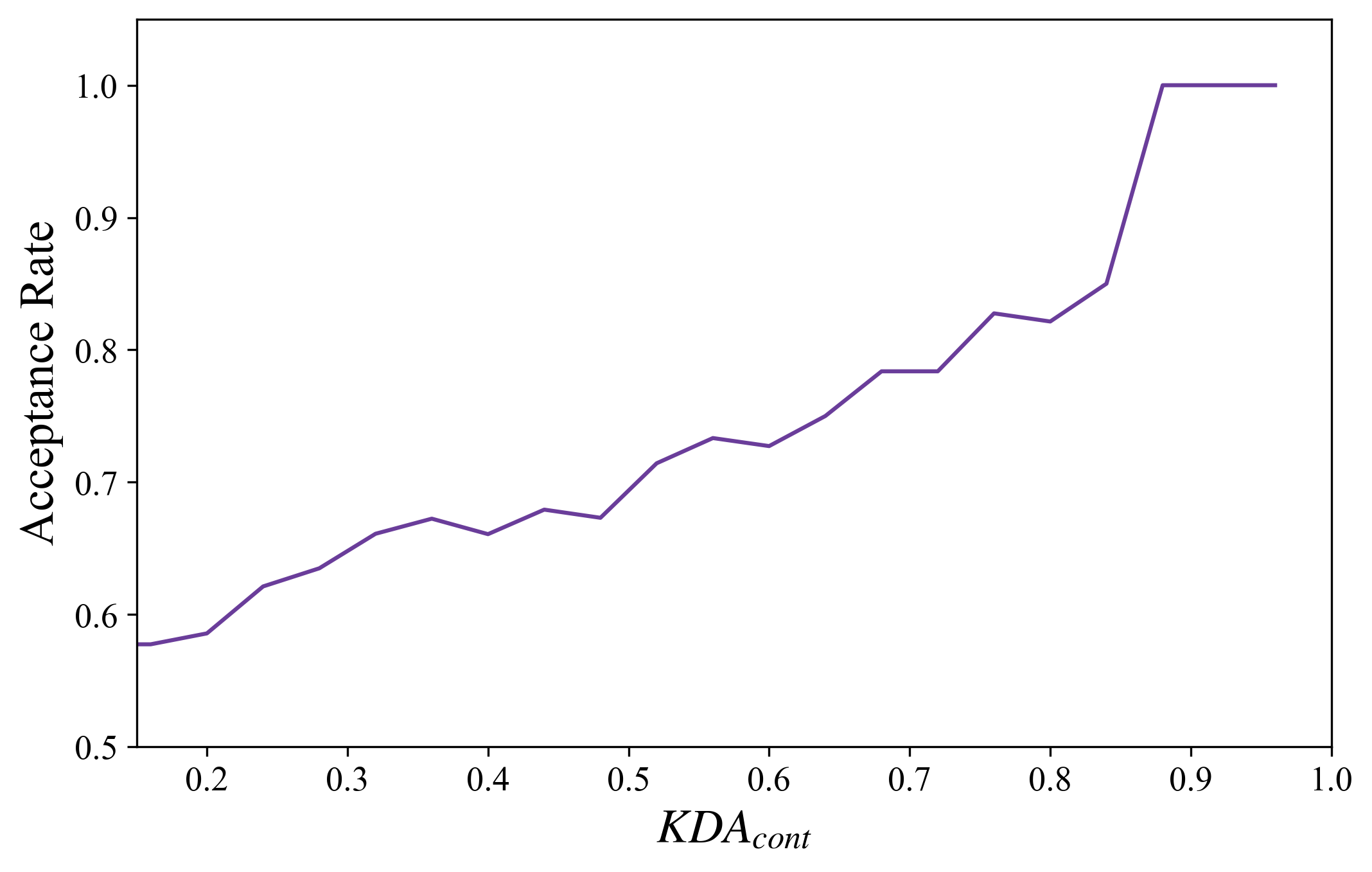}
    \caption{Cumulative graph showing how the acceptance rate of questions above a specific \ourssoft~value changes. Given \ourssoft~= 0.2, y-axis value is the ratio of question that got accepted among questions with \ourssoft~over 0.2.  61\%, 51\%, 39\%, and 19\% of total questions has \ourssoft~over 0.6, 0.7, 0.8, and 0.9, respectively. Thus, there are enough samples for each bin of \ourssoft~threshold.}
    
    
    
  \label{fig:acceptance_rate}
\end{figure}

The results were obtained from a distribution in which a sufficient amount of samples existed even in a significant interval. 


\begin{figure}[t]
  \includegraphics[width=0.45\textwidth]{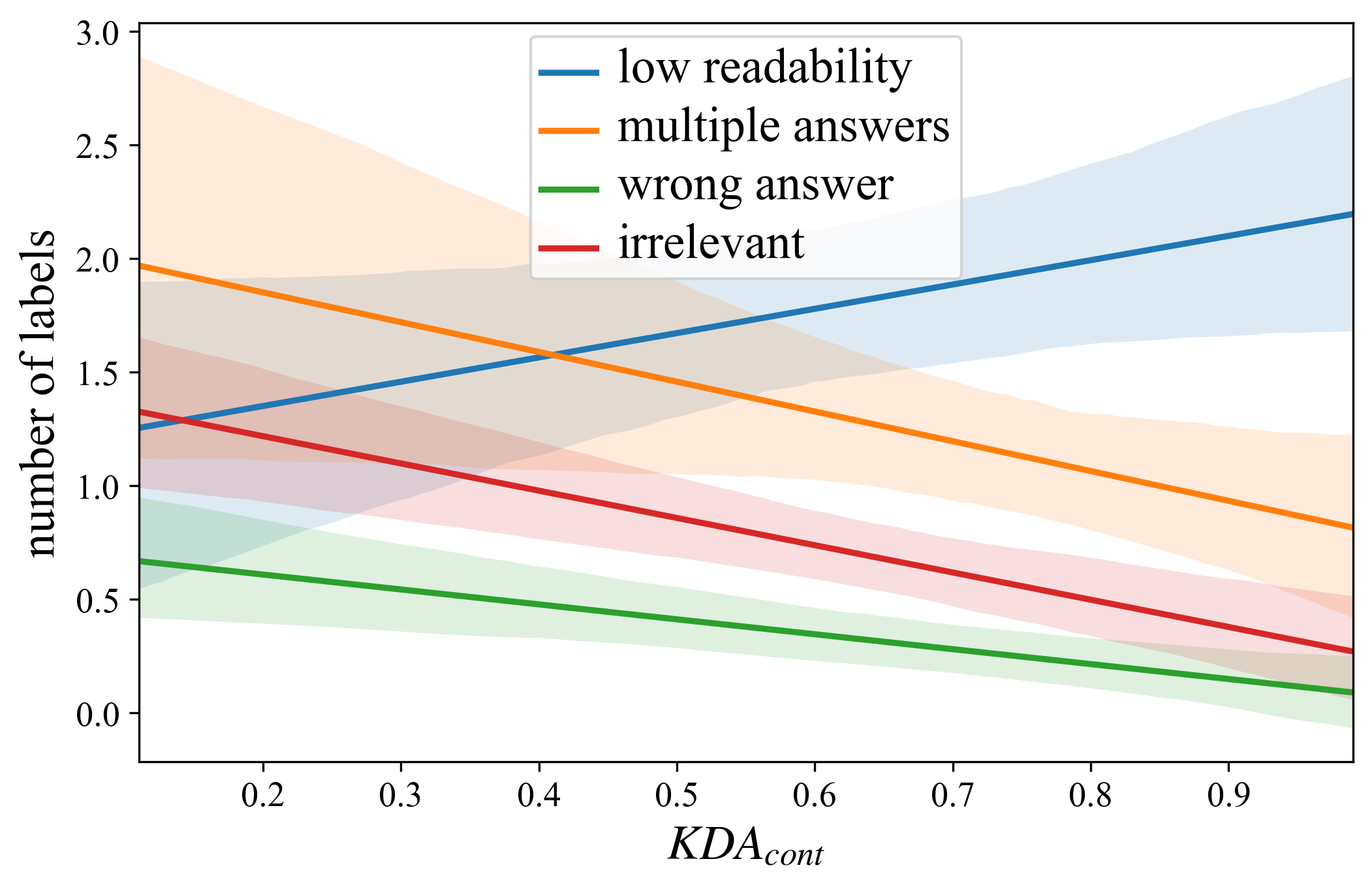}
    \caption{Regression graph of the number of expert labels by increasing \ourssoft~ of each question. The x-axis is \ourssoft, and the y-axis is the number of flaw labels responded to by experts for each question with specific \ourssoft~value. The light area corresponds to 95\% CI. }
  \label{fig:label_KDA_S}
\end{figure}

\begin{table}[t]
\begin{adjustbox}{width=1\columnwidth}
\begin{tabular}{@{}lccc@{}}
\toprule
            & \oursstar~  & Others+\oursstar~& Others \\ \midrule
Likert      & \underline{0.33}  & \textbf{0.42}       & 0.21   \\
Accept      & \underline{0.41}  & \textbf{0.49}       & 0.19   \\
Irrelevancy  & \underline{0.22}  & \textbf{0.23}        & -0.03   \\
Low Readability & \textbf{-0.17} & -0.19      & \textbf{-0.17}  \\
Multi-Ans.  & \underline{0.21}  & \textbf{0.35}       & 0.11    \\
Wrong Ans.  & \textbf{0.22}  & \underline{0.18}       & -0.18  \\ \bottomrule 
\end{tabular}
\end{adjustbox}
\caption{Test set Pearson Correlation from Random Forest classifier. Others represent n-gram based similarity metrics (BLEU, ROUGE, METEOR) and \oursstar~represent both \ourssoft~and \oursdisc. The results are averaged across 10 trials of 4-fold stratified cross-validation. Best result is marked in bold and second best result in underlined.}
\label{tab:randomforest} 
\end{table}

\begin{table*}[ht!]
\centering
\begin{adjustbox}{width=1\textwidth}
\begin{tabular}{clccccccccc}
\hline
\textbf{Examples} &
  \# &
  \textbf{Model} &
  \textbf{Dataset} &
  \textbf{Question} &
  \textbf{Fact} &
  \textbf{Options} &
  \textbf{\ourssoft} &
  \textbf{\oursdisc} &
  \textbf{Likert} &
  \textbf{BLEU} \\ \hline
\multirow{14}{*}{\begin{tabular}[c]{@{}c@{}}\ours\\ and\\ Experts\\ Agree\end{tabular}} &
  1 &
  Human &
  TabMCQ &
  \begin{tabular}[c]{@{}c@{}}Where can Coyoteite \\ be found?\end{tabular} &
  \begin{tabular}[c]{@{}c@{}}Coyoteite can be \\ found in rocks\end{tabular} &
  \begin{tabular}[c]{@{}c@{}}rocks \checkmark\\ scissors\\ spoons\\ pens\end{tabular} &
  0.90 &
  1.00 &
  3.29 &
  100 \\ \cline{2-11} 
 &
  1 &
  T5DG &
  TabMCQ &
  \begin{tabular}[c]{@{}c@{}}Where can Coyoteite \\ be found?\end{tabular} &
  \begin{tabular}[c]{@{}c@{}}Coyoteite can be\\ found in rocks\end{tabular} &
  \begin{tabular}[c]{@{}c@{}}rocks \checkmark\\ plastic\\ wood\\ glass\end{tabular} &
  0.90 &
  1.00 &
  3.71 &
  0 \\ \cline{2-11} 
 &
  2 &
  Human &
  TabMCQ &
  A cave is formed by \_. &
  \begin{tabular}[c]{@{}c@{}}A(n) cave is formed by\\ weathering\end{tabular} &
  \begin{tabular}[c]{@{}c@{}}weathering \checkmark\\ glacial erosion\\ plate tectonics\\ continental drift\end{tabular} &
  0.77 &
  0.87 &
  3.00 &
  100 \\ \cline{2-11} 
 &
  2 &
  QDG &
  TabMCQ &
  \begin{tabular}[c]{@{}c@{}}A cave is formed by \\ what process?\end{tabular} &
  \begin{tabular}[c]{@{}c@{}}A(n) cave is formed by\\ weathering\end{tabular} &
  \begin{tabular}[c]{@{}c@{}}weathering \checkmark\\ glaciers\\ volcanoes\\ erosion\end{tabular} &
  0.82 &
  1.00 &
  3.29 &
  62.5$^\dagger$ \\ \hline
\multirow{2}{*}{\begin{tabular}[c]{@{}c@{}}\ours~\\ and \\ Experts \\ Disagree\end{tabular}} &
  3 &
  \multicolumn{1}{l}{KDDG} &
  OBQA &
  \begin{tabular}[c]{@{}c@{}}Quartz crystals are\\ made up of\end{tabular} &
  \begin{tabular}[c]{@{}c@{}}a quartz is made of\\ six-sided transparent\\ crystals\end{tabular} &
  \begin{tabular}[c]{@{}c@{}}hexagons \checkmark\\ oval\\ square\\ sphere\end{tabular} &
  0.39 &
  0.38 &
  3.00 &
  0 \\ \cline{2-11} 
 &
  4 &
  \multicolumn{1}{l}{KDDG} &
  SciQ &
  \begin{tabular}[c]{@{}c@{}}Assume a molecule must \\ cross a plasma membrane \\ into what?\end{tabular} &
  \begin{tabular}[c]{@{}c@{}}Assume a molecule must \\ cross the plasma membrane \\ into a cell... (excerpt)\end{tabular} &
  \begin{tabular}[c]{@{}c@{}}cell \checkmark\\ electron\\ tissue\\ plasma\end{tabular} &
  0.81 &
  0.93 &
  1.71 &
  0 \\ \hline
\end{tabular}
\end{adjustbox}
\caption{Samples questions where \ours~agrees or disagrees with Expert Likert scale. $^\dagger$: For QGD, BLEU for question stem was evaluated.}
\label{tab: goodbadexamples}
\end{table*}

\subsection{RQ2: Expert Judgment on \oursstar}
\label{subsection:teacher}
\paragraph{Setup}
To see whether \ours~can measure the educational value of MCQs in a real-world classroom setting, we conducted a survey with previous or current secondary school science teachers. 
In particular, we randomly sampled 8 facts (32 questions) from each dataset (24 facts, 96 MCQs in total) among the facts that were selected for the human study. 
Given a pair of fact and an MCQ, we first asked the annotators to rate whether they would use the given MCQ to be used in a classroom to test the fact on a 4-point Likert scale (1: "Strong Reject", 2: "Weak Reject", 3: "Weak Accept", 4: "Strong Accept"), as shown in Table \ref{tab:response}. 
Whenever an annotator rated an MCQ a score of 1 or 2, we further asked them why the given MCQ was unsatisfactory. They were able to choose one option between low readability, multiple answers, wrong answer, irrelevancy, and other. More experimental details can be found in the Appendix. 

\paragraph{Results}
First, as shown in Figure~\ref{fig:acceptance_rate}, acceptance ratio increases as \ourssoft~increases. 
Notably, we can see that \textbf{82\%} of the questions which scored \ourssoft~over \textbf{0.8} were accepted to be used in a classroom setting. 
This shows that \ourssoft~can serve as a robust filtering metric to determine whether to use the generated question in a classroom setting or not. 

We further examined whether \ourssoft~can explain why an MCQ is unsatisfactory. 
Specifically, we fitted a linear regression model that takes \ourssoft~as an input, and predicts the MCQ quality measures (Low Readability, Multiple Answers, Wrong Answer, Irrelevance). 
As seen in Figure~\ref{fig:label_KDA_S}, MCQs with higher \ourssoft~tend to be more relevant, while they are less likely to have multiple answers or wrong answer. 
However, \ourssoft~does not show prediction power on Low Readability. 
We believe that low readability is particularly hard to capture with \ourssoft: for example, students may find an MCQ hard to understand due to jargons and difficult vocabulary, while language models do not have such problem. 

Finally, we examined which metrics best explain the given MCQ's quality annotated by experts—Likert scale and other binary labels of rejection such as Acceptance\footnote{Accept labels are given to questions with likert score higher than 2.5.}, Irrelevancy, Low Readability, Multiple Answers, and Wrong Answer. 
In particular, we trained a \textit{Random Forest classifier}, which predicts MCQ's various quality measures based on the given input metrics.  
As shown in Table~\ref{tab:randomforest}, using \oursdisc~and \ourssoft~outperforms using ngram-based similarity metrics (BLEU, ROUGE, METEOR) by a large margin for all but one quality measure of `Readability'.
Combining \ours~metrics with similarity-based metrics, they show a strong synergy, reaching correlation of \textbf{0.49} with Accept labels. 

\subsection{Case Study}
Here, we examine cases where \ourssoft~agrees with Expert Likert scores, aptly capturing the assessment quality of the question.
As shown in Table~\ref{tab: goodbadexamples}, BLEU cannot take into account novel, yet convincing distractors (e.g., ``wood'' from \#1) not present in the gold distractor set. 
However, \ourssoft~is able to take account the discriminative value of the distractors, and gives a high score to the MCQ, regardless of their resemblance to gold distractors. This is also demonstrated in \#2, where QDG-generated question scored higher than the human generated question in Likert, \ourssoft, \oursdisc, but not in $BLEU$.

We also present the cases where \ourssoft~fails to measure the educational value of the given question.

\textbf{Low \ours, High Likert: } \#3 shows the example of common types of question where \ourssoft~cannot be high. In order to get the question correctly, the student need to know that hexagon has six sides, which is not stated in the given fact. If the question requires some steps of reasoning, LMs would have relatively low correctness compared to the human solver, resulting in low \ourssoft~and \oursdisc. This can be resolved using LMs with better readig comprehension and reasoning abilities. 

\textbf{High \ours, Low Likert: } \#4 shows the case where \ourssoft~is high but Likert score is low. 
As shown in the table, the question asks the portion of the fact that is an assumption. While instructors can likely mark the question to have a low assessment value as it test a mere assumption, language model can not discern the quality of the fact the question is testing. Since SciQ is a crowd-source dataset, we noticed some questions don't assess meaningful target facts.

\section{Conclusion}

In this paper, we proposed Knowledge Dependent Answerability to measure the assessment value of the generated Multiple Choice Question. We formulated \ours~as the probability that a student will solve the question correctly if the student know the target fact being tested. Then, we proposed \oursdisc~and \ourssoft~to approximate \ours, treating PLMs as individual human solvers. Since both metrics are reference-free evaluation metrics, they can be applied without restrictions unlike previous metrics. They can be also applied as filters for question generation for educational use cases.

On 3 real-world MCQ datasets, \ourssoft~ and \oursdisc~demonstrated a high correlation with \ours~and the expert Likert score that measures the usability of the question as an assessment tool in a classroom setting. Notably, using \ourssoft~and \oursdisc~along with current n-gram based metrics drastically increased the overall correlation with these expert labels, as well as the prediction power of specific rejection reasons such as irrelevancy and wrong answer. We released a code and model weights to easily measure \ourssoft~and \oursdisc~for a given question and a fact pair. Future research may address expanding the metric's applicability to other types of assessment questions, such as short answer questions or multi-hop questions.


\section{Limitations}
\begin{table*}[t]
\centering
\begin{tabular}{@{}l|lll|lllll@{}}
\toprule
      & obqa & tabMCQ & sciQ & human & qg+dg & dg   & kddg & All  \\ \midrule
kappa & 0.18 & 0.07   & 0.31 & 0.09  & 0.23  & 0.24 & 0.22 & 0.20 \\ \bottomrule
\end{tabular}
\caption{Cohen's kappa coefficient for inter-annotator agreement. All 21 coefficient between 7 annotator were averaged.}
\label{tab:kappa}
\end{table*}

In this section, we discuss limitations of our methods and experiments. 



\subsection{Prompt Can Bias Solvers' Decision}
Our prompt-based method has limitations at estimating the student's knowledge on the fact, especially for multiple-answer situations. 
Suppose a question ``Select an option that is in a liquid state at 20 Celcius'' with multiple choices ``water, orange juice, desk, and air'', where the choice ``water'' is the only labeled answer, targeting a fact ``Water is a liquid at 20 Celcius.''
However, as both ``water'' and ``orange juice'' are valid answers for the question, prompting the target fact can critically bias students' decision towards the only labeled answer ``water,'' discounting the other valid answer ``orange juice''.
In such cases, though the questions are of low quality, \ours~cannot filter out the questions if the multiple answers are not labeled properly.

\subsection{Too Easy Questions}

Our experiments have an assumption that a well-designed MCQ is answerable if a student knows the target fact. 
However, there are edge cases where an MCQ question is answerable regardless of whether the student actually knew the fact. For instance, an answer can be easily found in the question stem itself or the distractors can be easily ruled out by simple topic irrelevance. 
In our experiment setting, this implies that there may have been students who are classified as knowledgeable, but didn't actually have the target knowledge as too easy questions cannot exactly distinguish the students' knowledge. 




\subsection{Difficulty of Measuring PLM's Ignorance} 
For PLM-based solvers, we do not have access to the huge training corpus in most cases, nor can we guarantee that the PLM \textit{knows} a fact contained in the training corpus. 
We leave this as future work, as language models with language comprehension ability but without any knowledge are needed for such measure. 

\subsection{Low Agreement between Teachers}
Table \ref{tab:kappa} shows the inter-rater agreement between two expert labelers as measured by Cohen's kappa. When we asked the teachers whether they would use the given MCQ in a real classroom setting (Section~\ref{subsection:teacher}), the inter-rater agreement was relatively low, showing 0.2 on average among 7 annotators. 
We believe this is because the annotators had different subjective views on what makes a ``good'' MCQ. However, we noticed that reviewers have higher agreement on ``bad'' MCQs: questions with \ourssoft~lower than 0.3 shows kappa coefficient over 0.3, and questions with \ourssoft~lower than 0.2 shows kappa coefficient over 0.4.

\subsection{Availability of PLMs for Low-Resource Languages}
Effectiveness of \ourssoft~and \oursdisc~has only been tested in English questions with PLMs trained with English corpus. Since the metric depends on using PLMs that are trained to have a good reading comprehension ability, the use of the metric might be limited for low-resource languages. 



\section{Ethical Considerations}

This study suggests that the question generation method can be applied in the real world, especially in the educational domain. Question for the purpose of assessment plays a vital role in the education system, and  automatic question generation can reduce the cost of education providers. In this process, we expect our study to contribute to reducing inequality by increasing educational opportunities. 


Despite such needs and efforts in AQG systems, automatically generated questions have shortcomings, even if they pass our proposed evaluation methodology. Since our method is designed to focus on knowledge dependency, several problems might remain in the filtered question, such as gender or racial bias \cite{hirota2022gender}.

Our study aims to evaluate the absolute value of the question through language models rather than the existing psychometric-based relative evaluation method. Since there hasn't been much discussion about AQG systems in general, considerable attention and additional research is required before deploying AQG systems in real-world and evaluating them with our proposed metric. 


\section*{Acknowledgements}

We thank all of our human experiment and expert experiment participants for their time. We also thank the anonymous reviewers for their valuable feedback.


\newpage
\bibliography{anthology,custom}
\bibliographystyle{acl_natbib}

\newpage
\appendix
\section{Appendix}
\label{sec:appendix}

\subsection{DataSet Preprocessing}

\noindent
\textbf{OBQA, SciQ} For OBQA and SciQ, we followed provided train-valid-test split. Questions containing "of above" or "of the above" are filtered out since they are exceptions to the problem formation. Train and valid splits are used to train the question generation model, and the questions we used in the human experiment and labeling are generated from the test set.

\noindent
\textbf{TabMCQ} Since TabMCQ does not provide splits, and we divided it into 6:1:1, train, valid, test, respectively. Since the raw tabMCQ dataset does not provide evidence of the fact as a natural language form, the fact is generated by concatenating the expressions on the table. Regent tables 27 to 43 were omitted because they were not written in natural language formations. Facts were filtered if several instances were concatenated inside one column. Like other dataset settings, the question generation model is trained using train validation splits, and the questions from the test split are used for the evaluation.

\subsection{Question Generator Training Details}

\noindent
\textbf{QG, DG} T5-large model was used, with max target token length of 512, target token length of 64, early stopping patience of 10 epochs, learning rate of 0.00001 with AdamW optimizer, and batch size of 24 (4 A100 machines in DDP, batch size of 6 for each machine). 

\noindent
\textbf{KDDG} implementation is done by description and open source of its original work. \cite{kddg} Since the source code and the resources are partially accessible and the appendix part, which is mentioned to contain the experiment settings, was missing, we re-implemented many parts of the code. 
Probase, which is renamed to Microsoft Concept Graph \footnote{\url{https://concept.research.microsoft.com/Home/Download}}, is used as the knowledge base. Latent Dirichlet Allocation(LDA) for topic modeling uses Gensim library\footnote{\url{https://radimrehurek.com/gensim/}} and is trained on the processed Wikipedia corpus provided by the library. Feature Extractor for distractor selector were implemented except omitted \emph{Contextual Embedding similarity} and \emph{Web-search Score} at the code. LambdaMart \footnote{\url{https://github.com/lezzago/LambdaMart}} is used as a Ranker, while the number of trees is 2, and the learning rate is 0.1.

\subsection{Human Experiment Details}
Among the generated questions, we filtered out incompletely generated questions that have less than 3 distractors. Among all the facts, we randomly sampled 40 facts from each dataset. As visualized in \ref{fig:main_figure}, students are asked to answer the question with and without the corresponding fact. Total 116 student joined the experiment questionnaire made by Typeform \footnote{\url{https://www.typeform.com/}}. We split the experiment group into 8 sessions, and each group takes 2 hours to solve the questions following the instructions. Due to COVID-19, we conducted the experiment remotely, while a non-face-to-face video call solution Zoom \footnote{\url{https://zoom.us/}}, the experiment was conducted by checking the participant's face and the participant's screen.
(fact validation)

In addition to the experiment setting described in the \ref{sec: experiment}, we asked the participants whether they knew the relevant facts as the prior knowledge after answering the questions without facts and before answering with facts. We showed the fact independently and let the students choose 'yes' or 'no' to answer the question, "Did you previously know the below fact?"
Its purpose was to resolve the limitation of the prompt-based knowledge-providing method and utilize the response as a gold label of the direct knowledge state of students. However, we can not find a meaningful explanation for the relationship between this label and our automatic metrics or human Likert score. 
 Considering the average correctness of TabMCQ questions with fact is 0.99, we can expect that TabMCQ has very high relevance between the given fact and the question. Surprisingly, over 33\% of students answered the question incorrectly but responded knowing the given fact and vice versa. Further research is needed to explain the difference between having knowledge and being presented with a prompt.

\subsection{Expert Labeling Details}
The experiment was conducted with 7 high school science teachers. During the 1 hour and 30 minute experiment, the teacher received a 150\$ amazon gift card as a reward. A total of 8 types of facts and 32 questions were presented for each dataset, and a questionnaire was conducted on a total of 96 questions. For each multiple-choice question, the questionnaire was asked whether to use the Likert scale in the actual educational field, and the question to choose the most prominent reason not to use the problem was answered.

\subsection{MCQ Solver implementation}
The entire model list are as follows.
\begin{itemize}
    \item T5-cbqa-small, T5-cbqa-large, T5-cbqa-xxl \cite{roberts2020much}
    \item bert-base, bert-large \cite{kenton2019bert}
    \item Roberta-base, Roberta-large \cite{liu2019roberta}
    \item MPNet \cite{song2020mpnet}
    \item SciBERT \cite{beltagy2019scibert}
    \item XLNet-base, XLNet-large \cite{yang2019xlnet}
    \item BioBERT-base, BioBERT-large \cite{lee2020biobert}
    \item DistillBERT-base, DistillRoberta-base \cite{Sanh2019DistilBERTAD}
    \item ALBERT-xl, ALBERT-xxl \cite{DBLP:journals/corr/abs-1909-11942}
    \item MatsciBERT \cite{gupta_matscibert_2022}
\end{itemize}
\label{MCQ Solving train}
Except for T5 models trained on closed-book question answering, all models are fine-tuned to the RACE dataset to answer MCQs. AdamW optimizer was used with a learning rate of 1e-5 for base models and 1e-6 for large models. Early stopping patience of 5 epochs was used. Base models were trained with a batch size of 32, and large models were trained with a batch size of 16. A cloud instance is used on google cloud service\footnote{\url{https://cloud.google.com/gcp}}  to train the solver models. Overall, the cloud instance cost for training is about to 14k\$, while the pricing policy is 'on demand,' which costs much higher than the preemptive instance.

For Table \ref{tab: N1}, used models are as follows.
\begin{itemize}
    \item LMs < 1GB (4 LMs): 
    DistillBert-base, DistillRoberta-base, Albert-xl, T5-cbqa-small
    \item LMs < 1.5GB (4 LMs): 
    bert-base, Roberta-base, XLNet-base, MPNet
    \item LMs < 1.5GB (11 LMs):
    DistillBert-base, DistillRoberta-base, Albert-xl, T5-cbqa-small, bert-base, Roberta-base, BioBERT-base, SciBERT, MatsciBERT, XLNet-base, MPNet
\end{itemize}

\subsection{Random Forest Implementation Detail}
\label{random forest}
Among tree depths of 2 to 4, depth 2 was selected as Others (BLEU, ROUGE, METEOR) showed the best performance in that depth. The test set was composed of 2 key facts per dataset (8 questions per dataset, total 24 questions). The result was averaged across 10 trials of 4-fold stratified cross-validation. No hyper-parameters were tuned from the default sklearn random forest setup.

\begin{table}[!t]
\centering
\label{tab:permodelkdascore}
\begin{adjustbox}{width=0.8\columnwidth}
\begin{tabular}{@{}lllll@{}}
\toprule
            & OBQA  & TabMCQ & SciQ   & All    \\\midrule
Human       & -0.57 & -0.18  & 0.06   & -0.01  \\
KDDG        & 0.01  & 0.13   & 0.75*  & 0.42*  \\
DG          & 0.59  & -0.01  & 0.65   & 0.61** \\
QDG       & 0.13  & 0.26   & 0.58   & 0.64** \\
DGen models & 0.32  & 0.16   & 0.65** & 0.51** \\ \bottomrule
\end{tabular}
\end{adjustbox}
\caption{Correlation of \ourssoft~and Human Likert Score per generate model. DGen models are counting both KDDG and DG models to figure the performance on evaluating distractor generation models.}
\end{table}

\begin{table}[!t]
\centering
\begin{tabular}{l|lll}
\toprule
       & avg. Rq & avg. Rf & avg. Rq+f \\ \hline
OBQA   & 0.58    & 0.80    & 0.71      \\
TabMCQ & 0.53    & 0.42    & 0.99      \\
SciQ   & 0.56    & 0.42    & 0.96     \\ \bottomrule
\end{tabular}
\caption{Average Correctness of Datasets. TabMCQ and SciQ are notably higher than OBQA.}
\end{table}

\begin{table}[t!]
\centering
\begin{tabular}{@{}cccc@{}}
\toprule
Sub Metric & \begin{tabular}[c]{@{}c@{}}Model Count\\ ( Total Size )\end{tabular} & \begin{tabular}[c]{@{}c@{}}\ours\\ ( Valid )\end{tabular} & \begin{tabular}[c]{@{}c@{}}Likert\\ ( Test )\end{tabular} \\ \midrule
\ourssmall & 4 (3.5GB)                                                            & 0.740                                                 & 0.377                                                   \\
\ourslarge & 10 (19.2GB)                                                          & 0.784                                                 & 0.421  \\ \hline                                            
\end{tabular}
\caption{ Two sub metrics of \ourssoft~developed for the convenience of use. \ourssmall~uses T5-cbqa-small, ALbert-xl, MPNet, SciBert to calculate \ourssoft~, and \ourslarge~uses T5-cbqa-small, T5-cbqa-large, ALbert-xl, MPNet, SciBert, bert-base, BioBert-base, Roberta-base, Roberta-large, XLNet-large.}
\label{tab: sub metric}
\end{table}
\subsection{Sub Metrics}
Since utilizing all models used to calculate \ourssoft~or \oursdisc~needs huge compute resources, we provide two variant metrics, \ourssmall~and \ourslarge.  The subset of entire models is selected under the constraints of the number and total sizes of models. We used questions without an expert label as a validation set to pick a combination of the highest correlation to \ourssoft and measure correlation to an expert Likert score as a test, and correlation is at Table \ref{tab: sub metric}.


\begin{table*}[]
\centering
\begin{adjustbox}{width=0.8\textwidth}
\begin{tabular}{@{}ccccc@{}}
\toprule
Examples      & Eg1                                                                                            & Eg2                                                                                                                          & Eg3                                                                                                                                  & Eg4                                                                                                                                  \\ \midrule
Dataset       & OBQA                                                                                           & OBQA                                                                                                                         & OBQA                                                                                                                                 & OBQA                                                                                                                                 \\ \midrule
QG model      & KDDG                                                                                           & KDDG                                                                                                                         & KDDG                                                                                                                                 & DG                                                                                                                                   \\ \midrule
Fact          & \begin{tabular}[c]{@{}c@{}}a beach ball \\ contains gas\end{tabular}                           & \begin{tabular}[c]{@{}c@{}}water is in the \\ solid state, called ice , \\ for temperatures\\ between 0 and 0 F\end{tabular} & \begin{tabular}[c]{@{}c@{}}friction acts to \\ counter the motion of \\ two objects when their \\ surfaces are touching\end{tabular} & \begin{tabular}[c]{@{}c@{}}friction acts to \\ counter the motion of\\  two objects when their\\  surfaces are touching\end{tabular} \\ \midrule
Question      & \begin{tabular}[c]{@{}c@{}}Which would you \\ likely find inside \\ a beach ball?\end{tabular} & \begin{tabular}[c]{@{}c@{}}Global warming is\\  lowering the world's \\ amount of\end{tabular}                               & \begin{tabular}[c]{@{}c@{}}When it's flying, \\ a plane has no \\ friction with the\end{tabular}                                     & \begin{tabular}[c]{@{}c@{}}When it's flying, \\ a plane has no \\ friction with the\end{tabular}                                     \\ \midrule
answer        & air                                                                                            & ice                                                                                                                          & ground                                                                                                                               & ground                                                                                                                               \\ \midrule
options       & \begin{tabular}[c]{@{}c@{}}food\\ gas\\ water\end{tabular}                                     & \begin{tabular}[c]{@{}c@{}}snow\\ water\\ air\end{tabular}                                                                   & \begin{tabular}[c]{@{}c@{}}power\\ air\\ water\end{tabular}                                                                          & \begin{tabular}[c]{@{}c@{}}air\\  water\\  sky\end{tabular}                                                                          \\ \midrule
gold\_options & \begin{tabular}[c]{@{}c@{}}steam\\ water\\ cheese\end{tabular}                                 & \begin{tabular}[c]{@{}c@{}}hurricanes\\ carbon dioxide\\ ocean levels\end{tabular}                                           & \begin{tabular}[c]{@{}c@{}}wings\\ clouds\\ air\end{tabular}                                                                         & \begin{tabular}[c]{@{}c@{}}wings\\ clouds\\ air\end{tabular}                                                                         \\ \midrule
Likert        & 2.57                                                                                           & 2.57                                                                                                                         & 2.71                                                                                                                                 & 3.0                                                                                                                                  \\
\ourssoft        & 0.11                                                                                           & 0.38                                                                                                                         & 0.27                                                                                                                                 & 0.29                                                                                                                                 \\
\oursdisc        & 0.08                                                                                           & 0.36                                                                                                                         & 0.40                                                                                                                                 & 0.54                                                                                                                                 \\
BLEU          & 33.3                                                                                           & 0.00                                                                                                                         & 33.3                                                                                                                                 & 33.3                                                                                                                                 \\ \bottomrule
\end{tabular}
\end{adjustbox}
\caption{ We report all cases of \ourssoft~$< $ 0.4 and $\text{human}_\text{likert}$ $>$ 2.5 }
\label{KDA_low}
\end{table*}

\begin{table*}[t]
\label{goodex}
\centering
\begin{adjustbox}{width=0.8\textwidth}
\begin{tabular}{@{}ccccc@{}}
\toprule
Examples      & Eg1                                                                             & Eg2                                                                           & Eg3                                                                                            & Eg4                                                                             \\ \midrule
Dataset       & TabMCQ                                                                          & TabMCQ                                                                        & TabMCQ                                                                                         & TabMCQ                                                                          \\ \midrule
QG model      & QG+DG                                                                           & QG+DG                                                                         & DG                                                                                             & DG                                                                              \\ \midrule
Fact          & \begin{tabular}[c]{@{}c@{}}water is an insulator\\  of electricity\end{tabular} & \begin{tabular}[c]{@{}c@{}}air is an insulator\\  of electricity\end{tabular} & \begin{tabular}[c]{@{}c@{}}warm is a term\\  that can describe \\ air temperature\end{tabular} & \begin{tabular}[c]{@{}c@{}}water is an insulator\\  of electricity\end{tabular} \\ \midrule
Question      & \begin{tabular}[c]{@{}c@{}}Water is an insulator\\  of what?\end{tabular}       & \begin{tabular}[c]{@{}c@{}}Air is an insulator\\  of what?\end{tabular}       & \begin{tabular}[c]{@{}c@{}}What does the term\\  warm describe?\end{tabular}                   & \begin{tabular}[c]{@{}c@{}}Water is an insulator\\  of what?\end{tabular}       \\ \midrule
answer        & electricity                                                                     & electricity                                                                   & air temperature                                                                                & electricity                                                                     \\ \midrule
options       & \begin{tabular}[c]{@{}c@{}}cold\\  heat\\  warmth\end{tabular}                  & \begin{tabular}[c]{@{}c@{}}heat\\  cold\\  warmth\end{tabular}                & \begin{tabular}[c]{@{}c@{}}precipitation\\  wind speed\\  cloud cover\end{tabular}             & \begin{tabular}[c]{@{}c@{}}heat\\  warmth\\  cold\end{tabular}                  \\ \midrule
gold\_options & \begin{tabular}[c]{@{}c@{}}wind\\ air\\ heat\end{tabular}                       & \begin{tabular}[c]{@{}c@{}}heat\\ water\\ metal\end{tabular}                  & \begin{tabular}[c]{@{}c@{}}wind speed\\ air pressure\\ optical phenomenon\end{tabular}         & \begin{tabular}[c]{@{}c@{}}wind\\ air\\ heat\end{tabular}                       \\ \midrule
Likert        & 2.14                                                                            & 2.42                                                                          & 2.29                                                                                           & 2.00                                                                            \\
\ourssoft        & 0.84                                                                            & 0.82                                                                          & 0.84                                                                                           & 0.84                                                                            \\
\oursdisc        & 0.93                                                                            & 0.93                                                                          & 0.86                                                                                           & 0.93                                                                            \\
BLEU          & 33.33                                                                           & 33.33                                                                         & 33.33                                                                                          & 33.33                                                                           \\ \bottomrule
\end{tabular}
\end{adjustbox}
\caption{ We report all cases of \ourssoft $> $ 0.8 and $\text{human}_\text{likert}$ < 2.5}
\end{table*}




\end{document}